\documentclass{article}
\PassOptionsToPackage{numbers,sort&compress}{natbib}
\usepackage[preprint]{corl_2026} 
\usepackage{float}
\usepackage{caption}
\usepackage{amsmath,amssymb,amsfonts}
\usepackage{bbm}
\usepackage{graphicx}
\usepackage{booktabs}
\usepackage{multirow}
\usepackage{enumitem}
\usepackage{algorithm}
\usepackage{algorithmic}
\usepackage{xcolor}
\usepackage{colortbl}
\definecolor{ResultShade}{HTML}{F7F7F7}
\definecolor{ResultAccent}{HTML}{8B1E1E}
\definecolor{ResultSD}{HTML}{737373}
\newcommand{\resultstat}[2]{\mbox{$#1${\small\color{ResultSD}$\,\pm\,#2$}}}
\newcommand{\resultbest}[2]{\mbox{\textcolor{ResultAccent}{\textbf{#1}}{\small\color{ResultSD}$\,\pm\,#2$}}}
\newcommand{\resultmethod}[1]{\textcolor{black}{\textbf{#1}}}
\usepackage{url}
\usepackage{hyperref}
\usepackage{wrapfig}
\usepackage{needspace}
\usepackage{ragged2e}
\newsavebox{\gaitphasebox}
\newsavebox{\simresultsbox}
\usepackage{tabularx}
\usepackage{changepage}
\usepackage[numbers,sort&compress]{natbib}

\usepackage{amsmath}
\usepackage{amssymb}

\newcommand{\ours}{TactileStep}

\usepackage{graphicx}
\usepackage{wrapfig}

\title{TactileStep: Sole Tactile Learning for Regulating Foot-Terrain Interaction in Humanoid Locomotion}

\author{
  \textbf{Zizhuo Wang\textsuperscript{*}, Ming-Ju Lee\textsuperscript{*}, }
  \textbf{Shaoting Zhu, Haozhe Lou,
  Hang Zhao\textsuperscript{$\dagger$},
  Yiming Li\textsuperscript{$\dagger$}}\\
  Tsinghua University\\
  {\small
    \textsuperscript{*}Equal contribution.
    \quad
    \textsuperscript{$\dagger$}Corresponding authors.
  }\\
  Website: \url{https://tactilestep.github.io/}
}

\begin{document}

\maketitle
\vspace{-1.0em}

\noindent\makebox[\textwidth][c]{%
    \includegraphics[width=0.92\textwidth]{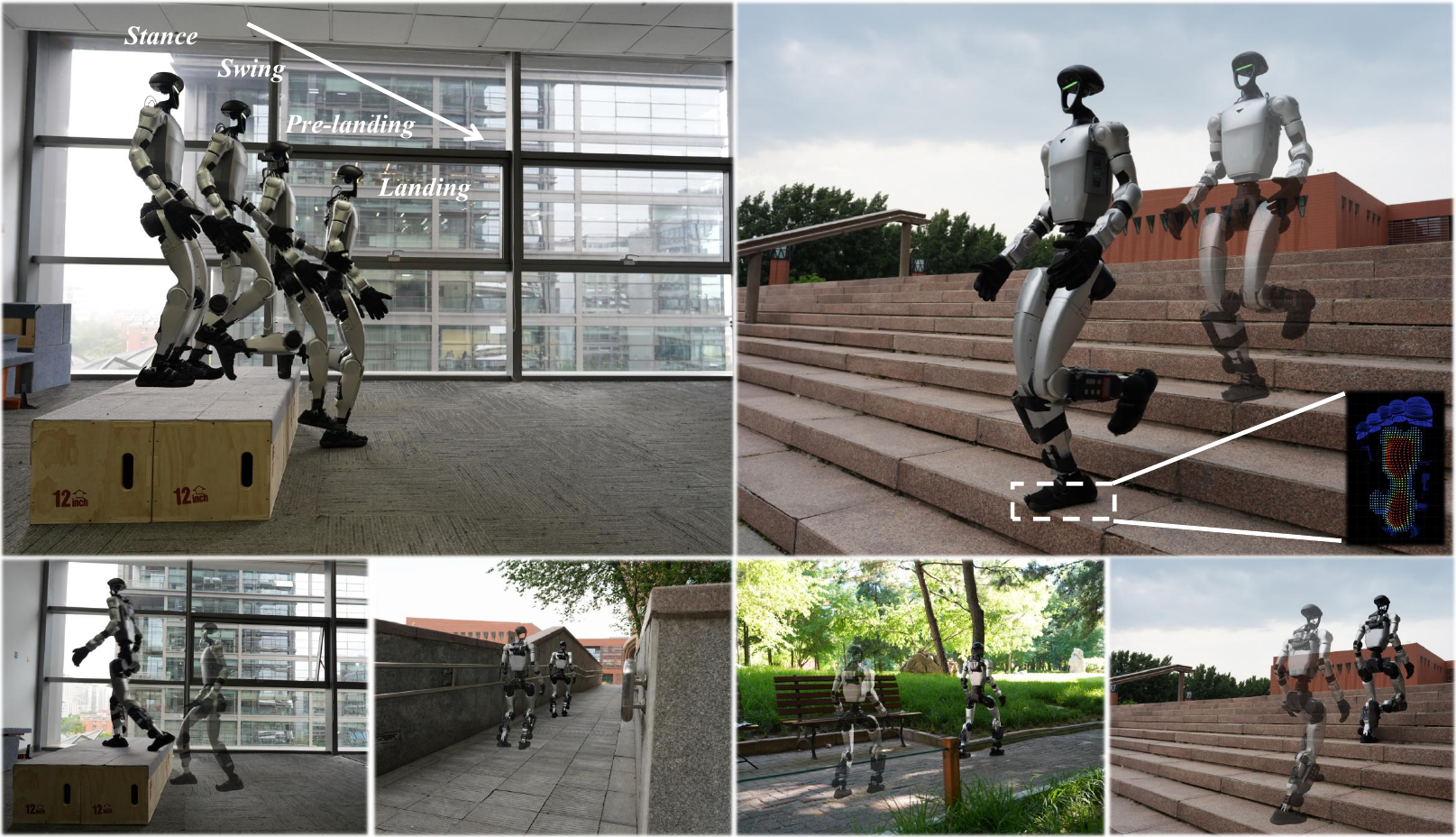}%
}
\vspace{-0.6em}

\begin{center}
\captionof{figure}{
\textbf{\ours{} turns foot--terrain contact from a passive outcome into a controllable signal.}
Humanoid parkour policies may traverse complex terrain while still landing harshly or forming fragile stance contacts.
We equip the robot with deployable sole-pressure feedback and train the policy to regulate how each foot lands, loads, and supports the body.
The figure shows real-world deployment across diverse terrains together with the four contact phases used by \ours{} to organize tactile regulation: swing, pre-landing, landing, and stance.
}
\end{center}
\vspace{-1.0em}

\begin{abstract}
Humanoid parkour policies can traverse various terrains, but task completion may mask challenges of harsh landings, edge contacts, and unstable stance contacts.
Humans naturally regulate foot–terrain interaction through tactile feedback, modulating contact compliance according to terrain stiffness. This highlights a key domain gap between humans and humanoid robots: the absence of rich tactile sensing in most humanoid systems.   
We address this problem with \textbf{\textit{\ours{}}}, a deployable tactile learning framework that brings sole pressure sensing into humanoid locomotion control for softer touchdowns and more stable support. \textbf{\textit{\ours{}}} aligns tactile simulation with the real pressure insole, allowing the policy to learn from the same contact features available on hardware.
During training, we use tactile and motion cues to recognize different foot-contact phases and apply phase-aware rewards that encourage safer landing and more stable stance.
Evaluated in simulation and on a Unitree G1 humanoid across diverse terrains, \ours{} reduces peak touchdown force by up to \textbf{48.8\%} and peak A-weighted impact noise by up to \textbf{30.1~dB} over a strong perceptive baseline, while increasing stance contact area by up to \textbf{23.8\%}.
\end{abstract}
\section{Introduction}
Humanoid locomotion has advanced rapidly, progressing from robust walking to perceptive terrain traversal and agile parkour~\cite{rl_bipedal,humanoidlocomotion_rl,learning_smooth,hugwbc,hwangbo2019learning}. Recent learning-based controllers can handle diverse, complex terrains using onboard perception and sim-to-real transfer~\cite{DWL,hpl,beamdojo,hikinginthewild,rpl,php,rudin2022learning}. However, a policy may complete a parkour task while landing with high impact, stepping near an edge, or relying on unstable support. Since stability, durability, and deployability are all essential in robotic applications, the next bottleneck for humanoid parkour is no longer only whether the robot can traverse an obstacle, but also whether it can establish safe, gentle, and stable contact while doing so.

This problem is most evident around each step. At touchdown, impact transients can induce vibration and generate noise, limiting long-term use in human-centered environments~\cite{quietwalk,quadruped_noise}. On stairs and platforms, a feasible foot placement may still yield bad support near edges~\cite{hikinginthewild,ttt-parkour,beamdojo}. During stance, small contact area or an offset center of pressure (CoP) can reduce the effective support margin. For humanoids, these local contact errors can quickly propagate into whole-body instability.

Existing sensing and learning pipelines therefore suffer from a sensing mismatch. Vision, height maps, and terrain reconstruction describe terrain geometry before contact~\cite{hikinginthewild,ttt-parkour,rpl,DWL,php,beamdojo,kumar2021rma,miki2022learning}, whereas soft landing and support stability depend on the contact state formed after touchdown. Although proprioception and estimated ground reaction forces (GRFs) provide impact-related feedback~\cite{quietwalk,grf_estimation,quadruped_noise}, they still leave the deployed policy without direct access to the required sole-pressure state. This gap naturally motivates a comparison with humans: humans rely on rich plantar pressure feedback to regulate landing impact, redistribute support, and recover balance after contact~\cite{hohne2011effects}. Tactile sensing provides useful guidance for humanoid controllers to achieve not only successful traversal, but also safer, gentler, and more stable contact.

Our key idea is that \textbf{{\textit{sole tactile sensing can turn the contact state under the foot into a deployable policy observation}}}. Because of the large sim-to-real gap in high-dimensional raw tactile data, we use a sole pressure array to measure how normal load is distributed over the foot and summarize it into compact features, such as normal force, contact area, and center of pressure (CoP). These features reveal not only contact timing, but also touchdown intensity and the quality of subsequent support. Rather than treating such information as privileged information, we use a hardware-available tactile representation to make contact quality part of closed-loop control.


We propose \textbf{\textit{\ours{}}}, a learning framework that makes sole tactile feedback part of humanoid parkour control. \textbf{\textit{\ours{}}} uses thin pressure insoles on a Unitree G1 humanoid robot and builds a compact tactile representation shared by simulation and hardware. At inference, the policy combines tactile features with proprioception and depth, allowing it to respond to landing impact and stance support. To organize contact objectives, we divide each foot motion into four phases: \textsc{Swing}, \textsc{Pre-Landing}, \textsc{Landing}, and \textsc{Stance}. Phase-conditioned rewards apply each objective where it is physically meaningful: limiting downward foot motion before contact, regulating impact at touchdown, and stabilizing support during stance. A dual-critic design separates dense and sparse reward groups according to their temporal structure. In summary, our contributions are threefold.
\begin{itemize}[leftmargin=1.5em]
    \item We introduce a \textbf{deployable sole-pressure representation}, aligning simulated and real tactile signals into normal force, contact area, and CoP features observed by the policy at deployment.
    
    \item We propose a \textbf{phase-aware tactile learning framework} that assigns contact-quality rewards to physically meaningful step segments, regulating touchdown impact and stance support.
    
    \item We validate \textbf{\ours{} on a Unitree G1} with onboard pressure insoles, demonstrating lower touchdown impact and improved tactile support metrics on different terrains.
\end{itemize}

\section{Related Work}
\label{sec:related_work}
\paragraph{Foothold Objectives and Support Regulation in Humanoid Locomotion.}
Recent perceptive humanoid policies improve foothold safety by converting terrain geometry into objectives for swing foot placement, including terrain edge detection, foot volume point penalties, overlap between the sole and terrain, and sampled foothold rewards~\cite{hikinginthewild,rpl,beamdojo}. 
These objectives serve as effective geometric proxies for foothold feasibility and reduce risky edge contacts during terrain traversal. 
However, for humanoids, foothold feasibility is only a prerequisite: stable locomotion also depends on the support formed after touchdown. 
A feasible foothold can still lead to fragile stance when sole contact is limited or the CoP approaches the support boundary. 
Prior studies show that plantar contact information supports support-region estimation, balance stabilization, and locomotion control under partial or uncertain contacts~\cite{partial_footholds_area,plantar_tactile,multicontact,tactile_balance,vukobratovic2004zero, romano2016static}. 
Building on these observations, our learning framework uses features computed from sole pressure to regulate both foothold acquisition and stance support.

\paragraph{Tactile Simulation and Sim2Real Learning.}
Prior tactile sim-to-real methods range from binary contact sensing \cite{yin2023rotating} and taxel-level representation learning \cite{yang2023tacgnn} to richer models of normal, shear, and distributed tactile responses \cite{yin2025learning,lin2025locotouch,wang2022tacto,si2022taxim}. 
This range highlights a practical trade-off: simpler abstractions ease scalable learning and transfer, while richer models retain more contact information at higher simulation complexity.
We balance the two for humanoid plantar interaction by distributing rigid-body contact forces over sole taxels and extracting normal force, contact area, and CoP. Both policy learning and sim-to-real alignment operate on these features, rather than the full spatial distribution of plantar pressure, retaining locomotion-relevant contact information while reducing modeling complexity and the tactile sim-to-real gap.
\paragraph{Soft Landing and Contact Regulation in Legged Locomotion.}
Soft landing is critical for legged robots operating near humans, since foot--ground impacts induce vibration, acoustic noise, and hardware wear~\cite{quietwalk,quadruped_noise,quietwalk_for_dogs,quietpaw,proprioceptive,landing_control,preemptive}. 
For bipedal robots, classical methods reduce landing impact through compliant hardware or local contact control, but often depend on specific mechanical designs and control assumptions~\cite{proprioceptive,landing_control,prelanding_control,preemptive}. 
Recent learning-based work offers a more scalable route, mainly in quadrupeds, by penalizing foot contact velocity, tuning gait and control parameters, or adapting policies to noise constraints~\cite{quietwalk_for_dogs,quadruped_noise,quietpaw}. 
Humanoid soft landing remains largely underexplored: QuietWalk learns a proprioceptive GRF predictor from plantar sensor data and penalizes predicted impact during training, but it leaves deployment-time sole feedback unused and is limited to regular walking~\cite{quietwalk}. 
Our work instead exposes contact features to the policy and uses phase-conditioned contact rewards to regulate landing impact on parkour terrains such as stairs and high platforms.

\section{Method}
\label{sec:method}
\vspace{-2mm}
\begin{figure}[t]
    \centering
    \makebox[\textwidth][c]{%
        \includegraphics[width=1\textwidth]{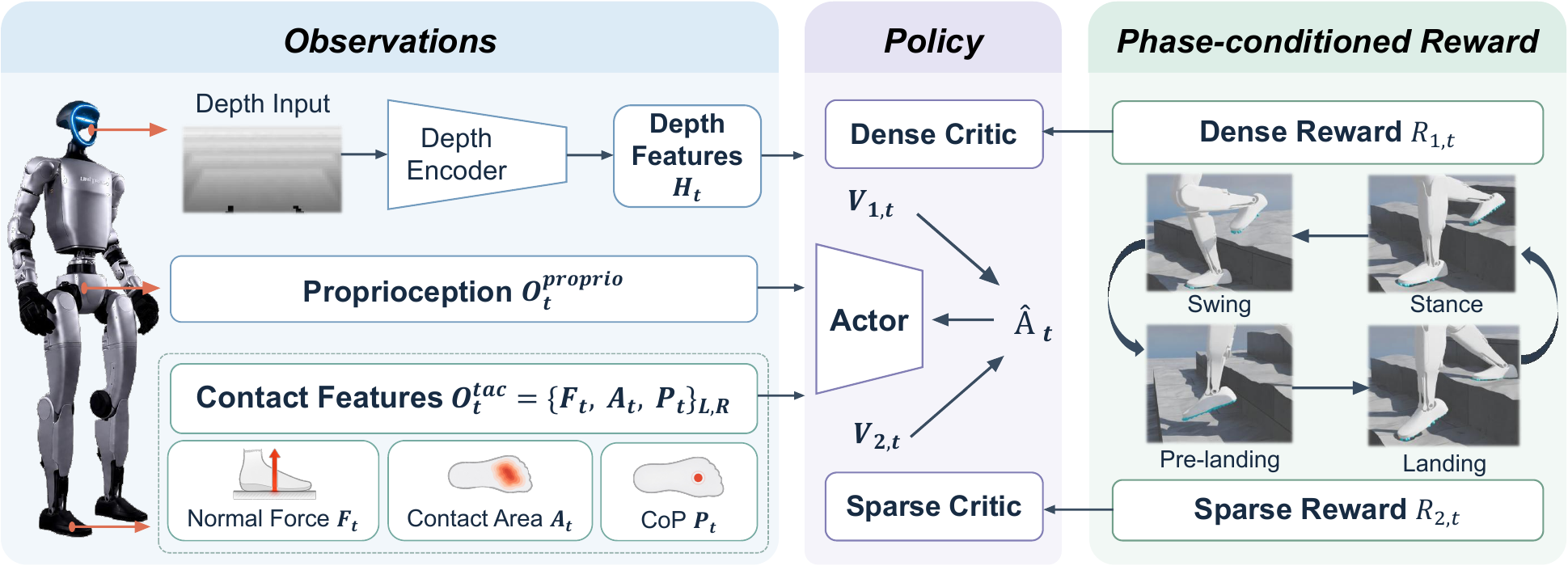}%
    }
    \vspace{1mm}
    \caption{
    \textbf{Overview of \ours{}.}
Our central design is to turn sole pressure into a deployable contact-state representation for policy learning. 
We first construct a lightweight tactile simulator that maps rigid foot--terrain contacts to a pressure array, from which contact features are extracted and exposed to the policy during both training and deployment.
Gait phases are estimated online and phase-conditioned rewards are designed to regulate touchdown impact and stance support.}
    \vspace{1mm}
    \label{fig:method_framework}
\end{figure}

We avoid exposing the actor to simulator-specific contact signals that do not transfer reliably to hardware. Instead, simulated contacts are mapped to deployable tactile features shared across simulation and hardware, while other privileged quantities are used only by the critics and reward computation.
\subsection{Problem Formulation}
\label{subsec:problem_formulation}
We formulate the control problem as a Partially Observable Markov Decision Process (POMDP) and use the Proximal Policy Optimization (PPO) ~\cite{schulman2017proximal} to
optimize the locomotion policy.
\paragraph{Observation Space.}
For each foot $f\in\{L,R\}$, the tactile observation at time $t$ is
$
\mathbf{x}^{f}_t =
\left[
\bar{F}^{f}_t,\ 
\mathbf{p}^{\mathrm{cop},f}_t,\ 
\bar{A}^{f}_t
\right],
$
where $\bar{F}^{f}_t$, $\mathbf{p}^{\mathrm{cop},f}_t\in\mathbb{R}^{2}$, and $\bar{A}^{f}_t$ denote the normalized normal force, CoP, and normalized contact-area ratio.
The actor observes:
\begin{equation}
\mathbf{o}^{a}_{t}
=
\underbrace{
\left\{
\boldsymbol{\omega}_{i},
\mathbf{g}_{i},
\mathbf{c}_{i},
\mathbf{q}_{i},
\dot{\mathbf{q}}_{i},
\mathbf{a}_{i-1}
\right\}_{i=t-h_p+1}^{t}
}_{\text{proprioceptive history}}
\oplus
\underbrace{
\left\{
\mathbf{x}^{L}_{j},
\mathbf{x}^{R}_{j}
\right\}_{j=t-h_{\mathrm{tac}}+1}^{t}
}_{\text{tactile history}}
\oplus
\mathcal{H}_{t},
\end{equation}
where the proprioceptive terms denote base angular velocity, projected gravity, velocity command, joint states, and previous action. $\mathcal{H}_{t}$ is the depth-observation history following~\cite{hikinginthewild}.
Inspired by~\cite{beamdojo,robotkeyframing,xu2023composite,huang2022reward}, we use two critics with the same privileged observation to manage different reward groups:
$
\mathbf{o}^{c_1}_{t}
=
\mathbf{o}^{c_2}_{t}
=
\mathbf{o}^{a}_{t}
\oplus
\left\{
v^{L}_{z,j},
v^{R}_{z,j},
\mathbf{e}^{L}_{j},
\mathbf{e}^{R}_{j}
\right\}_{j=t-h_{\mathrm{tac}}+1}^{t},
$
where $v^{f}_{z,j}$ is the vertical velocity of foot $f$, and $\mathbf{e}^{f}_{j}\in\{0,1\}^{4}$ is the one-hot encoding of its gait phase: swing, pre-landing, landing or stance.

\paragraph{Action Space.}
The policy outputs the target joint positions $\mathbf{a}_{t} \in \mathbb{R}^{29}$. Then, the joint torques $\boldsymbol{\tau}_{t}$ are computed via PD control:
$
\boldsymbol{\tau}_{t}
=
\mathbf{k}_{p}(\mathbf{a}_{t} - \mathbf{q}_{t})
-
\mathbf{k}_{d}\dot{\mathbf{q}}_{t}.
$
The specific PD gain values are adopted from~\cite{beyondmimic}. These computed torques are then applied to the actuators to execute the desired motion.
\paragraph{Reward Functions.}
The reward consists of task, regularization, safety, and adversarial motion prior terms:
$
r_t
=
r_{\mathrm{task},t}
+
r_{\mathrm{reg},t}
+
r_{\mathrm{safe},t}
+
r_{\mathrm{amp},t}.
$
These terms encourage goal-directed locomotion, suppress unsafe or inefficient motions, enforce joint limits, and promote natural motion styles~\cite{peng2021amp}. 
For dual-critic training, we organize the same reward into dense terms and sparse terms:
\begin{equation}
r_t
=
\underbrace{\sum_{m\in\mathcal{D}} w_m r_{m,t}}_{r^{\mathrm{dense}}_t}
+
\underbrace{\sum_{m\in\mathcal{S}} w_m r_{m,t}}_{r^{\mathrm{sparse}}_t}.
\end{equation}
The decomposition follows the temporal density of the learning signal:
dense terms provide frequent, continuous per-step feedback, whereas sparse terms become informative
only at discrete events, gates, or constraint violations. The two critics estimate the corresponding
returns, and their advantages are mixed for actor updates; the full reward grouping is given in
Appendix~\ref{app:reward_design}.

\subsection{Tactile Simulation}
\label{subsec:tactile_sim}
\begin{wrapfigure}{r}{0.43\columnwidth}
\vspace{-0.7em}
\centering
\includegraphics[
    width=\linewidth,
    trim=5 5 5 5,
    clip
]{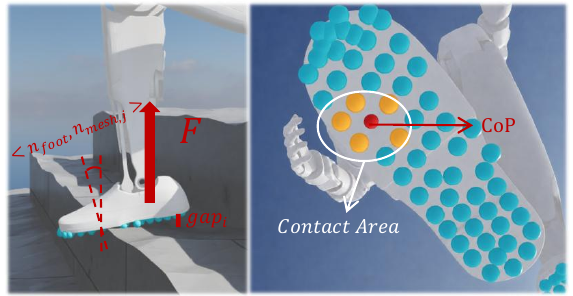}
\vspace{-0.5em}
\caption{\textbf{Tactile simulation.} 
Raycasting estimates taxel--terrain gaps and terrain normals for force distribution. The resulting pressure map provides the contact area and CoP for tactile feature extraction.}
\vspace{-2mm}
\end{wrapfigure}

We introduce a lightweight tactile simulator in Isaac Sim to generate compact sole-pressure observations from rigid-body foot--terrain contacts. 
Rather than simulating soft-body deformation, the method approximates the plantar pressure distribution by distributing the resultant foot contact force over a set of sole taxels. 
This design is computationally efficient for large-scale RL while retaining the contact features needed for tactile reward shaping and sim-to-real alignment.
The tactile model consists of three stages: force distribution, spatial diffusion, and feature extraction.

\paragraph{Force Distribution.}
For each foot, we place $M=60$ taxels on the sole. 
For taxel $i$, a raycast along the local sole normal estimates the nearest foot--terrain distance $\mathrm{gap}_i$ and the corresponding terrain normal $\mathbf{n}_{\mathrm{mesh},i}$. 
Let $\mathbf{n}_{\mathrm{foot}}$ denote the local sole normal. 
We compute an alignment- and distance-aware unnormalized weight

\begin{equation}
\tilde{w}_i
=
\left[
\operatorname{clip}
\left(
\frac{
\left\langle \mathbf{n}_{\mathrm{foot}}, \mathbf{n}_{\mathrm{mesh},i} \right\rangle
-
\eta_n
}{
1-\eta_n
},
0,
1
\right)
\right]^{\beta}
\exp
\left(
-\frac{\mathrm{gap}_i - \mathrm{gap}_{\min}}{\tau}
\right),
\quad
\tau =
\frac{\mathrm{gap}_{\max} - \mathrm{gap}_{\min}}{\ln 50}
\label{eq}
\end{equation}

where $\eta_n$ is the minimum normal-alignment threshold, 
$\mathrm{gap}_{\min}$ and $\mathrm{gap}_{\max}$ are the minimum and maximum taxel gaps. Given the resultant normal contact force $F$, the initial taxel force is
\begin{equation}
f_i
=
\frac{\tilde{w}_i}{\sum_{j=1}^{M}\tilde{w}_j}
F,
\label{eq:tactile_distribution}
\end{equation}
This produces an initial estimate of the sole-force distribution that is biased toward taxels that are close to the terrain and whose local contact normal is well aligned with the sole normal.

\paragraph{Spatial Diffusion.}
Rigid-body contact solvers may produce sparse or unstable point contacts. 
To approximate load spreading through the compliant sole and suppress isolated single-taxel activations, we diffuse each taxel force over its $k$ nearest neighbors:
\begin{equation}
f'_i
=
(1-\alpha) f_i
+
\alpha
\frac{1}{k}
\sum_{j \in \mathcal{N}(i)} f_j,
\label{eq:tactile_diffusion}
\end{equation}
where $\mathcal{N}(i)$ is the $k$-nearest-neighbor set of taxel $i$, and $\alpha\in[0,1]$ controls the diffusion strength. 
The resulting smoothed pressure map is
\begin{equation}
\mathcal{P}
=
\left\{
(x_i,y_i,f'_i)
\right\}_{i=1}^{M},
\label{eq:pressure_map}
\end{equation}
where $\mathbf{p}_i=(x_i,y_i)$ denotes the predefined position of taxel $i$ in the sole frame.

\paragraph{Feature Extraction.}
Raw taxel-level pressure maps are sensitive to sensor layout, calibration, and manufacturing differences. 
We therefore extract compact physical features that can be estimated by a wide range of plantar sensing systems. 
From the smoothed pressure map $\mathcal{P}$, we compute the resultant tactile force, contact-area ratio, and center of pressure:
\begin{equation}
F_{\mathrm{tac}}
=
\sum_{i=1}^{M} f'_i,
\qquad
A_{\mathrm{tac}}
=
\frac{1}{M}
\sum_{i=1}^{M}
\mathbb{I}(f'_i > \epsilon_f),
\qquad
\mathbf{p}_{\mathrm{cop}}
=
\frac{
\sum_{i=1}^{M} f'_i \mathbf{p}_i
}{
\sum_{i=1}^{M} f'_i + \epsilon
}.
\label{eq:tactile_features}
\end{equation}
Here $\epsilon_f$ is the taxel activation threshold, and $\epsilon$ prevents division by zero. 
The force and contact-area quantities are normalized before being passed to the policy, yielding $\bar{F}_{t}^{f}$ and $\bar{A}_{t}^{f}$. 
Compared with the full pressure map, these low-dimensional features are less dependent on a specific sensor configuration, which improves their suitability for sim-to-real transfer.

\paragraph{Sim2Real Alignment.}
\begin{wrapfigure}{r}{0.65\textwidth}
\centering
\includegraphics[width=\linewidth]{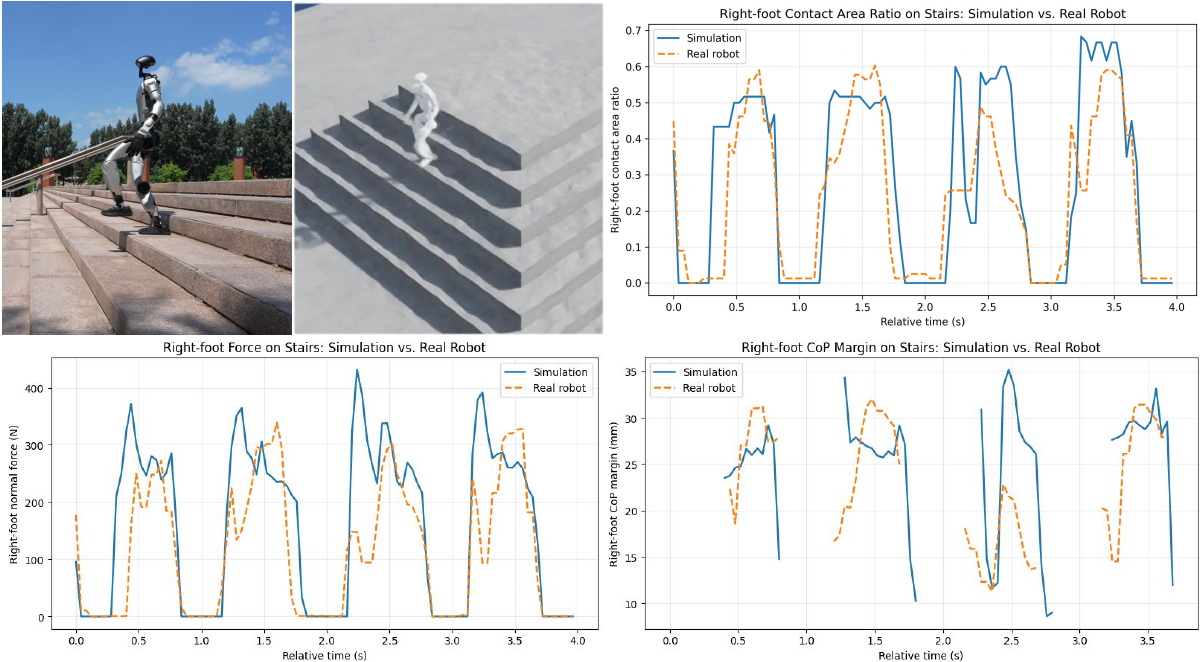}
\caption{\textbf{Feature-level tactile sim-to-real comparison during stair descent.}
Normal force, contact-area ratio, and CoP-margin traces from simulation and hardware are obtained
with the tactile-independent Baseline policy.}
\label{fig:sim2real_alignment}
\end{wrapfigure}
We assess sim2real alignment using the tactile-independent Baseline policy in both domains, avoiding the coupling between tactile feedback and policy's behavior.
Figure~\ref{fig:sim2real_alignment} compares $F_{\mathrm{tac}}$, $A_{\mathrm{tac}}$, and $\rho_{\mathrm{cop}}$ over a short stair-descent segment.
The simulated and measured traces exhibit similar ranges and trends despite residual sim-to-real differences.
For hardware calibration, an MLP maps raw sensor readings to force using manufacturer-provided data.
\par
\WFclear


\begin{lrbox}{\gaitphasebox}
\begin{minipage}{\textwidth}
\begin{minipage}[t]{0.54\textwidth}
\vspace{0pt}
\RaggedRight
\textbf{Gait Phase Estimation.}\quad
Many humanoid locomotion controllers use a fixed periodic gait clock to assign each foot to either \textsc{Swing} or \textsc{Stance}~\cite{wococo,Hold_my_beer,peng2025gait}. 
This representation is sufficient for regular walking, but becomes too coarse for regulating foot--terrain interaction. We therefore use a four-phase gait representation that is more detailed and physically meaningful for each foot:
$
s_t^f \in 
\{Swing, PreLanding,Landing,Stance\},\quad f \in \{L,R\},
$
where \textsc{PreLanding} and \textsc{Landing} capture the short transition windows around touchdown.
\end{minipage}\hfill
\begin{minipage}[t]{0.42\textwidth}
\vspace{0pt}
\scriptsize
\captionsetup{font=scriptsize,labelfont=bf,singlelinecheck=false,skip=3pt,hypcap=false}
\hrule
\captionof{algorithm}{Gait Phase Inference}
\label{alg:gait_phase_estimation}
\hrule
\smallskip
\noindent\textbf{Input:} $F,A,v_z,h,c^{-},\ell^{-}$ \\
\noindent\textbf{Output:} $s,c,\ell$
\smallskip
\begin{algorithmic}[1]
\STATE $c \leftarrow (F>F_{\rm th}) \lor (A>A_{\rm th})$
\STATE $p \leftarrow (v_z<0) \land (h<h_{\rm pre})$
\IF{$\neg c$}
    \STATE $\ell \leftarrow 0$
    \STATE $s \leftarrow \textsc{PreLanding}$ if $p$ else $\textsc{Swing}$
\ELSE
    \STATE $\ell \leftarrow N_{\rm land}$ if $\neg c^{-}$ else $\max(\ell^{-}-1,0)$
    \STATE $s \leftarrow \textsc{Landing}$ if $\ell>0$ else $\textsc{Stance}$
\ENDIF
\RETURN $s,c,\ell$
\end{algorithmic}
\smallskip
\hrule
\end{minipage}
\end{minipage}
\end{lrbox}
\Needspace{\dimexpr\ht\gaitphasebox+\dp\gaitphasebox+4\baselineskip\relax}
\subsection{Phase-Aware Rewards for Soft Landing and Support Stability}
\label{subsec:phase_rewards}
\noindent\usebox{\gaitphasebox}\par
\medskip


The estimated gait phase is used to route tactile rewards to the portions of the gait cycle in which they are physically relevant. 
Let
$
\mathbf{1}^{f}_{\phi,t}
=
\mathbb{I}(s_t^f=\phi)
$
indicate whether foot $f$ is in phase $\phi$ at time $t$. 
The phase-aware tactile reward is defined as
\begin{equation}
r^{\mathrm{tac}}_t
=
\sum_{f\in\{L,R\}}
\sum_{\phi\in\Phi}
\mathbf{1}^{f}_{\phi,t}
r^{f}_{\phi,t}, \quad
\Phi
=
\{
\textsc{Swing},
\textsc{PreLanding},
\textsc{Landing},
\textsc{Stance}
\}.
\label{eq:phase_reward}
\end{equation}
This formulation prevents touchdown penalties from being applied during normal swing and prevents stance-stability terms from being evaluated before contact is established.

\paragraph{Soft-Landing Reward.}
Soft landing is encouraged by regulating both the pre-contact approach and the post-contact impact transient. 
During \textsc{PreLanding}, we penalize aggressive downward motion:
\begin{equation}
r^{f}_{\mathrm{pre},t}
=
-
w_v
\left[
\max(0,-v^{f}_{z,t})
\right]^2
-
w_a
\left(a^{f}_{z,t}\right)^2 ,
\label{eq:prelanding_reward}
\end{equation}
where $v^{f}_{z,t}$ and $a^{f}_{z,t}$ are the vertical velocity and acceleration of foot $f$. 
This term discourages the foot from approaching the terrain with excessive downward speed or acceleration.

During \textsc{Landing}, tactile feedback is used to penalize large impact forces and rapid increases in contact load:
\begin{equation}
r^{f}_{\mathrm{land},t}
=
-
w_F
\left(\bar{F}^{f}_{t}\right)^2
-
w_{\Delta F}
\left[
\max(0,\Delta \bar{F}^{f}_{t})
\right]^2 ,
\label{eq:landing_reward}
\end{equation}
where
$
\Delta \bar{F}^{f}_{t}
=
\bar{F}^{f}_{t}
-
\bar{F}^{f}_{t-1}.
$ To further suppress impulsive touchdown events, we also penalize the peak force and peak force growth within the landing window $\mathcal{W}^{f}_{\mathrm{land}}$:
\begin{equation}
r^{f}_{\mathrm{peak},t}
=
-
w^{\mathrm{pk}}_F
\max_{\tau\in\mathcal{W}^{f}_{\mathrm{land}}}
\left(\bar{F}^{f}_{\tau}\right)^2
-
w^{\mathrm{pk}}_{\Delta F}
\max_{\tau\in\mathcal{W}^{f}_{\mathrm{land}}}
\left[
\max(0,\Delta \bar{F}^{f}_{\tau})
\right]^2 .
\label{eq:peak_landing_reward}
\end{equation}
Together, these terms encourage softer and smoother touchdowns.

\paragraph{Support-Stability Reward.}
During \textsc{Stance}, the reward encourages broad, centered, and temporally stable support:
\begin{equation}
r^{f}_{\mathrm{stance},t}
=
w_A \bar{A}^{f}_{t}
+
w_{\rho}\rho^{f}_{\mathrm{cop},t}
-
w_{\Delta p}
\left\|
\mathbf{p}^{\mathrm{cop},f}_{t}
-
\mathbf{p}^{\mathrm{cop},f}_{t-1}
\right\|_2^2 .
\label{eq:stance_reward}
\end{equation}
Here, $\bar{A}^{f}_{t}$ rewards a larger contact area, 
$\rho^{f}_{\mathrm{cop},t}$ denotes the center-of-pressure margin to the support boundary, and the final term penalizes abrupt center-of-pressure shifts. 
This reward discourages partial edge contacts and unstable pressure migration, which are especially detrimental on discontinuous or uneven terrain.

Overall, the phase-aware tactile rewards shape the policy toward cautious foot placement before contact, low-impact touchdown during landing, and stable load distribution during stance.

\section{Experimental Results}
\label{sec:experiments}
We evaluate \ours{} in both simulation and real-world environments through three questions:
\begin{itemize}[leftmargin=1.5em]
\item \textbf{Q1:} Does \ours{} produce \textbf{softer and quieter touchdowns}?
\item \textbf{Q2:} Does \ours{} establish more \textbf{stable foot--terrain support}?
\item \textbf{Q3:} Which \textbf{tactile observations and rewards} drive the improvements?
\end{itemize}

\subsection{Experimental Setup} 
\label{subsec:exp_setup} 
We train all policies in NVIDIA Isaac Sim with Isaac Lab~\cite{nvidia2025isaaclab}, using parallel simulation for RL. Training is performed on an NVIDIA RTX 4090 GPU and parallelized over 2048 humanoid agents. We use the 29-DoF Unitree G1 robot for both simulation training and physical deployment.
Simulation evaluation uses 4,096 trials per policy and terrain, with one episode per parallel environment (Appendix~\ref{app:sim_eval_protocol}).
Hardware evaluation uses 20 samples per condition.

\paragraph{Baseline and Ablations.}
\begin{adjustwidth}{1.2em}{0pt}
\noindent\textbf{External baseline.}
We compare \ours{} with the perceptive parkour policy from Hiking in the Wild~\cite{hikinginthewild}, which serves as a strong vision-based baseline for agile terrain traversal.

\noindent\textbf{w/o tac obs.}
This variant removes online sole-pressure features from the policy input, evaluating whether tactile observations provide useful contact cues beyond visual and proprioceptive inputs.

\noindent\textbf{w/o soft landing.}
This variant removes the soft-landing rewards, isolating the contribution of tactile feedback in reducing touchdown impact and regulating the pre-contact landing process.

\noindent\textbf{w/o stable.}
This variant removes the stable-support rewards, evaluating the role of tactile feedback in encouraging complete sole support and maintaining stable contact after landing.
\end{adjustwidth}

\begin{figure}[htbp]
    \centering
    \vspace{2mm}
    \includegraphics[width=0.96\textwidth]{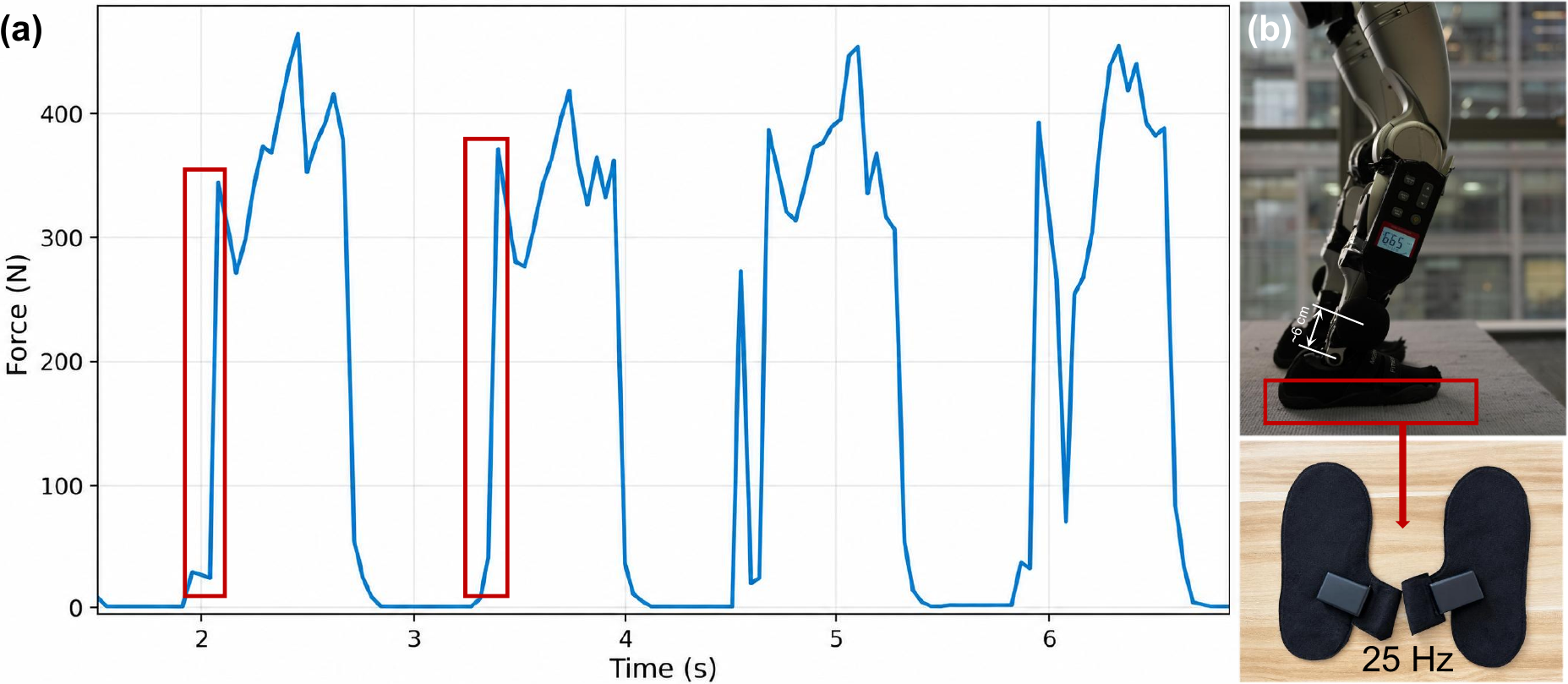}
    \vspace{3mm}
    \caption{(a) Left-foot force curve during stair ascent with \ours{}.
The red box indicates the first touchdown rising edge used to compute the hardware landing impact force $F_{\mathrm{impact}}$.
(b) The sound level meter is rigidly mounted on the lateral lower leg, facing downward,
approximately 6 cm above the lowest joint; the tactile insole is installed beneath the foot.}
    \label{fig:force}
\end{figure}

\paragraph{Metrics.}
We evaluate foot--terrain interaction quality mainly with three metrics.

\begin{adjustwidth}{1.2em}{0pt}
\noindent\textbf{Landing impact force} $F_{\mathrm{impact}}$ measures the transient normal force at touchdown. In simulation, it is computed as the maximum normal force within the landing window. On hardware, it is extracted from the first rising edge of the force curve after touchdown as shown in Figure~\ref{fig:force}.

\noindent\textbf{Contact area ratio} $A_c$ is measured during stance, with larger values indicating more complete sole support and better utilization of the valid support region.

\noindent\textbf{Peak acoustic noise} $L_{A,\mathrm{peak}}$ is reported in real-world experiments as the peak A-weighted sound level induced by foot--ground impact and indirectly reflects the magnitude of the landing impact.The effective background noise floor remained approximately 55--60 dB across hardware tests and
was dominated by the G1's onboard fan rather than ambient environmental noise.
\end{adjustwidth}

\subsection{Simulation Results} \label{subsec:sim_eval}

\newcommand{\best}[1]{\textbf{#1}}

\begin{table}[H]
\centering
\caption{
\textbf{Simulation evaluation} of support quality on terrains.
}
\label{tab:sim_support_quality}
\setlength{\tabcolsep}{4.2pt}
\renewcommand{\arraystretch}{1.1}
\resizebox{\textwidth}{!}{
\begin{tabular}{l >{\columncolor{ResultShade}}c c c >{\columncolor{ResultShade}}c c c >{\columncolor{ResultShade}}c c c >{\columncolor{ResultShade}}c c c}
\toprule
\multirow{2}{*}{Metric}
& \multicolumn{3}{c}{\textbf{Flat}}
& \multicolumn{3}{c}{\textbf{Slope}}
& \multicolumn{3}{c}{\textbf{Upstairs}}
& \multicolumn{3}{c}{\textbf{Downstairs}} \\
\cmidrule(lr){2-4} \cmidrule(lr){5-7} \cmidrule(lr){8-10} \cmidrule(lr){11-13}
& \resultmethod{\ours{}} & \textit{w/o stable} & Baseline
& \resultmethod{\ours{}} & \textit{w/o stable} & Baseline
& \resultmethod{\ours{}} & \textit{w/o stable} & Baseline
& \resultmethod{\ours{}} & \textit{w/o stable} & Baseline \\
\midrule
Contact Area Ratio $\uparrow$
& \resultbest{0.929}{0.054} & \resultstat{0.750}{0.068} & \resultstat{0.912}{0.020}
& \resultbest{0.805}{0.044} & \resultstat{0.624}{0.051} & \resultstat{0.804}{0.023}
& \resultbest{0.536}{0.042} & \resultstat{0.462}{0.040} & \resultstat{0.480}{0.012}
& \resultbest{0.545}{0.051} & \resultstat{0.455}{0.048} & \resultstat{0.487}{0.017} \\

CoP Margin [mm] $\uparrow$
& \resultbest{35.79}{0.54} & \resultstat{29.90}{2.34} & \resultstat{35.21}{0.14}
& \resultbest{34.24}{0.37} & \resultstat{27.55}{1.72} & \resultstat{34.20}{0.11}
& \resultbest{29.05}{1.16} & \resultstat{25.56}{1.35} & \resultstat{26.03}{0.40}
& \resultbest{29.43}{1.02} & \resultstat{25.91}{1.20} & \resultstat{27.07}{0.36} \\
\bottomrule
\end{tabular}
\vspace{1mm}
}
\end{table}
\begin{lrbox}{\simresultsbox}
\begin{minipage}{\textwidth}
\begin{minipage}[t]{0.53\textwidth}
\vspace{0pt}
Figure~\ref{fig:sim_impact} first examines touchdown impact in simulation, corresponding to \textbf{Q1}. Across all tested terrains, \ours{} yields the lowest landing impact force, with a clear margin over the Hiking baseline. More importantly, removing the soft-landing rewards consistently increases the impact force, which provides the ablation evidence for \textbf{Q3}: the phase-aware soft-landing objective is the key factor behind the reduced touchdown transients.
\par\smallskip
Table~\ref{tab:sim_support_quality} evaluates stance support quality, corresponding to \textbf{Q2}. 
Across the four terrains, \ours{} consistently improves both contact area and CoP margin over the \textit{w/o stable} variant, and is generally better than the baseline. 
The gains over the baseline are modest on flat ground and slopes, likely because these relatively simple and continuous terrains already allow the baseline to form stable support, leaving limited room for further improvement. 
Larger gains emerge on stairs, where foothold geometry imposes more challenging support conditions.
\end{minipage}\hfill
\begin{minipage}[t]{0.45\textwidth}
\vspace{0pt}
\centering
\includegraphics[width=\linewidth]{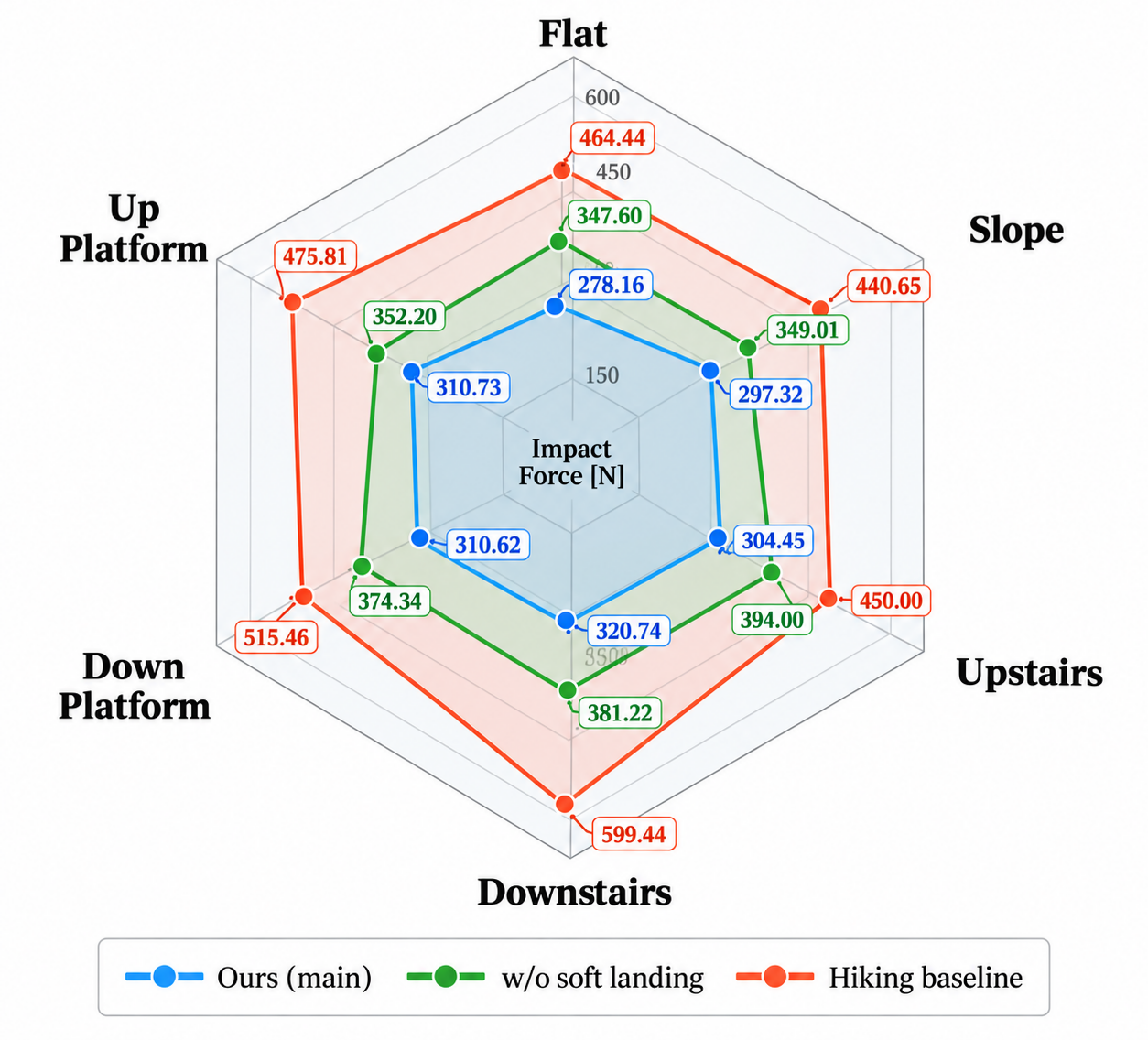}
\captionsetup{hypcap=false}
\captionof{figure}{\textbf{Simulation evaluation} of landing impact force on diverse terrains.}
\label{fig:sim_impact}
\end{minipage}
\end{minipage}
\end{lrbox}
\Needspace{\dimexpr\ht\simresultsbox+\dp\simresultsbox+\baselineskip\relax}
\noindent\usebox{\simresultsbox}\par
\medskip
This ablation answers \textbf{Q3}, showing that the stable-support rewards improve how the foot settles after touchdown. Taken together, the simulation results show that \ours{} does not simply make the robot land softly; it also helps the robot establish broader and more centered stance contacts.

Under the same simulation protocol, Table~\ref{tab:sim_locomotion} shows that \ours{} achieves success rates at least as high as those of Baseline.
Velocity-tracking error and traversal time are generally slightly higher but comparable in magnitude.
Energy consumption is higher, possibly due to longer traversal times and more active joint regulation.

\begin{table}[htbp]
\centering
\caption{\textbf{Standard locomotion metrics} in simulation.}
\label{tab:sim_locomotion}
\setlength{\tabcolsep}{4pt}
\renewcommand{\arraystretch}{1.08}
\resizebox{\textwidth}{!}{
\begin{tabular}{lcc cc cc cc}
\toprule

\multirow{2}{*}{Terrain}
& \multicolumn{2}{c}{Success Rate}
& \multicolumn{2}{c}{Velocity RMSE [m/s]}
& \multicolumn{2}{c}{Traversal Time [s]}
& \multicolumn{2}{c}{Energy [J]} \\

\cmidrule(lr){2-3}
\cmidrule(lr){4-5}
\cmidrule(lr){6-7}
\cmidrule(lr){8-9}

& \resultmethod{\ours{}} & Baseline
& \resultmethod{\ours{}} & Baseline
& \resultmethod{\ours{}} & Baseline
& \resultmethod{\ours{}} & Baseline \\

\midrule

Stair up
& 100\% & 99.98\%
& \resultstat{0.24}{0.02} & \resultstat{0.21}{0.02}
& \resultstat{5.7}{0.4} & \resultstat{5.6}{0.3}
& \resultstat{1126.3}{66.6} & \resultstat{802.2}{97.0}
\\

Stair down
& 100\% & 99.98\%
& \resultstat{0.20}{0.02} & \resultstat{0.19}{0.03}
& \resultstat{6.1}{0.6} & \resultstat{6.0}{0.5}
& \resultstat{1232.9}{91.0} & \resultstat{887.4}{110.4}
\\

Platform up
& 100\% & 99.93\%
& \resultstat{0.24}{0.02} & \resultstat{0.23}{0.03}
& \resultstat{4.9}{0.5} & \resultstat{4.9}{0.6}
& \resultstat{939.8}{50.5} & \resultstat{867.1}{72.8}
\\

Platform down
& 100\% & 92.26\%
& \resultstat{0.22}{0.02} & \resultstat{0.21}{0.15}
& \resultstat{4.9}{0.6} & \resultstat{4.9}{0.8}
& \resultstat{1033.0}{73.6} & \resultstat{1005.3}{119.7}
\\

Flat
& 99.98\% & 99.19\%
& \resultstat{0.20}{0.08} & \resultstat{0.17}{0.03}
& \resultstat{6.5}{3.3} & \resultstat{5.6}{2.2}
& \resultstat{834.4}{165.4} & \resultstat{644.5}{76.1}
\\

Slope
& 99.63\% & 99.63\%
& \resultstat{0.21}{0.17} & \resultstat{0.19}{0.05}
& \resultstat{6.0}{2.8} & \resultstat{5.2}{2.1}
& \resultstat{995.7}{126.1} & \resultstat{786.3}{47.9}
\\

\bottomrule
\end{tabular}
}
\end{table}

\subsection{Real-World Results} \label{subsec:real_eval} 

We further deploy the learned policies on the real humanoid to examine whether the contact improvements transfer to hardware. Table~\ref{tab:real_world_results} summarizes tactile and acoustic measurements across different terrains. Since force, noise, and contact area depend on terrain geometry and material, we focus on controlled within-terrain comparisons.

The results first answer \textbf{Q1}. Across the tested terrains, \ours{} produces lower landing impact force than the Hiking baseline, with particularly clear gains on platform ascent and descent where height discontinuities induce strong touchdown transients. The same trend appears in acoustic measurements: \ours{} also yields lower peak A-weighted noise. The agreement between force and sound confirms that the deployed policy produces softer touchdowns on hardware.

The results also answer \textbf{Q2}. On terrains where sustained stance support is meaningful, \ours{} achieves larger contact areas than the Hiking baseline, indicating broader sole support after touchdown. Thus, the policy does not reduce impact by weakening contact; it improves the landing--support trade-off under the same terrain conditions.

Finally, the \textit{w/o tac. obs.} ablation addresses \textbf{Q3}. Although this variant often improves over Hiking, it remains weaker than \ours{} in key hardware measurements, especially acoustic noise and contact area. This confirms the central role of deployed sole-pressure observations: without online tactile feedback, the policy cannot reliably regulate the realized contact state after touchdown.

\begin{table*}[t]
\centering
\caption{
\textbf{Real-world results} across diverse terrains.
}
\label{tab:real_world_results}
\setlength{\tabcolsep}{4.0pt}
\renewcommand{\arraystretch}{1.08}
\resizebox{\textwidth}{!}{
\begin{tabular}{l c c >{\columncolor{ResultShade}}c c c >{\columncolor{ResultShade}}c c c >{\columncolor{ResultShade}}c}
\toprule
\multirow{2}{*}{Terrain}
& \multicolumn{3}{c}{$F_{\mathrm{impact}}\downarrow$ [N]}
& \multicolumn{3}{c}{$L_{\mathrm{A}}\downarrow$ [dB]}
& \multicolumn{3}{c}{$A_c\uparrow$} \\
\cmidrule(lr){2-4}
\cmidrule(lr){5-7}
\cmidrule(lr){8-10}
& Baseline & w/o tac. obs. & \resultmethod{\ours{}}
& Baseline & w/o tac. obs. & \resultmethod{\ours{}}
& Baseline & w/o tac. obs. & \resultmethod{\ours{}} \\
\midrule
Stair up
& \resultstat{499.9}{40.7} & \resultstat{375.2}{47.3} & \resultbest{349.8}{32.5}
& \resultstat{101.2}{3.7} & \resultstat{85.1}{3.4} & \resultbest{72.2}{2.8}
& \resultstat{0.470}{0.013} & \resultstat{0.421}{0.028} & \resultbest{0.482}{0.016} \\

Stair down
& \resultstat{359.3}{23.2} & \resultstat{353.3}{38.6} & \resultbest{344.1}{20.1}
& \resultstat{97.2}{1.2} & \resultstat{84.9}{3.9} & \resultbest{67.1}{1.4}
& \resultstat{0.483}{0.020} & \resultstat{0.417}{0.044} & \resultbest{0.598}{0.027} \\

Platform up
& \resultstat{695.0}{49.4} & \resultstat{371.5}{50.5} & \resultbest{355.7}{39.8}
& \resultstat{110.1}{6.0} & \resultstat{98.2}{5.3} & \resultbest{83.0}{4.1}
& -- & -- & -- \\

Platform down
& \resultstat{608.9}{38.6} & \resultstat{445.7}{54.1} & \resultbest{404.6}{43.2}
& \resultstat{90.8}{5.2} & \resultstat{87.5}{4.4} & \resultbest{75.9}{3.6}
& -- & -- & -- \\

Flat
& \resultstat{201.3}{23.3} & \resultstat{202.1}{32.4} & \resultbest{191.3}{17.7}
& \resultstat{90.6}{2.5} & \resultstat{85.9}{2.9} & \resultbest{66.5}{1.6}
& \resultstat{0.453}{0.011} & \resultstat{0.362}{0.035} & \resultbest{0.510}{0.014} \\

Slope
& \resultstat{255.0}{34.9} & \resultstat{204.7}{42.1} & \resultbest{202.0}{31.0}
& \resultstat{83.4}{1.0} & \resultstat{77.1}{2.0} & \resultbest{69.9}{1.1}
& \resultstat{0.486}{0.026} & \resultstat{0.490}{0.033} & \resultbest{0.499}{0.030} \\
\midrule
\end{tabular}
}
\vspace{0.5em}
\end{table*}

\section{Conclusion}
\label{sec:conclusion}
We introduced \ours{}, a sole-tactile learning framework that makes foot--terrain interaction a first-class objective in humanoid locomotion. Rather than treating terrain traversal as a binary success criterion, \ours{} equips the policy with deployable pressure-based contact features that reveal how each foot lands, loads, and supports the body. A calibrated tactile simulator aligns these features between simulation and hardware, while phase-aware rewards regulate contact quality at the proper moments of each step, from pre-landing to stance. Experiments in simulation and on a Unitree G1 humanoid show that \ours{} reduces touchdown impact and acoustic noise, improves stance support, and remains effective across diverse terrains. These results suggest that sole tactile feedback offers a practical path toward humanoid locomotion that is not only robust, but also gentler, quieter, and more physically aware.

\section{Limitations and Future Work}
TactileStep focuses on improving foot--terrain contact quality during parkour locomotion, but several limitations remain. The policy is trained and evaluated within a bounded command range, so its generalization to substantially faster motions, where contact duration is shorter and impact transients are larger, remains unexplored. Our gait phase estimator also relies on hand-designed contact and motion cues; learning phase representations directly from onboard observations may improve robustness across speeds and terrains. In addition, the tactile simulator assumes rigid foot--terrain contact, and its fidelity on deformable or granular surfaces has not been validated. Long-term sensor durability, drift, and recalibration of the pressure insole are also not evaluated. Finally, our current evaluation does not systematically characterize failure cases or recovery behavior. More broadly, quieter and lower-impact locomotion can benefit human-centered deployment and hardware longevity, while broader deployment will require validating these sensing and control assumptions over longer-term and more diverse conditions.


\clearpage
\acknowledgments{
We thank the anonymous reviewers and Area Chair for their constructive feedback and suggestions. This research was funded by KEYSTONE ELECTRICAL (ZHEJIANG) CO.}


\bibliography{example}  

\clearpage
\appendix

\section{Additional Results and Ablations}
\subsection{Evaluation Protocol in Simulation}
\label{app:sim_eval_protocol}

\begin{figure*}[h]
    \centering
    \includegraphics[width=\textwidth]{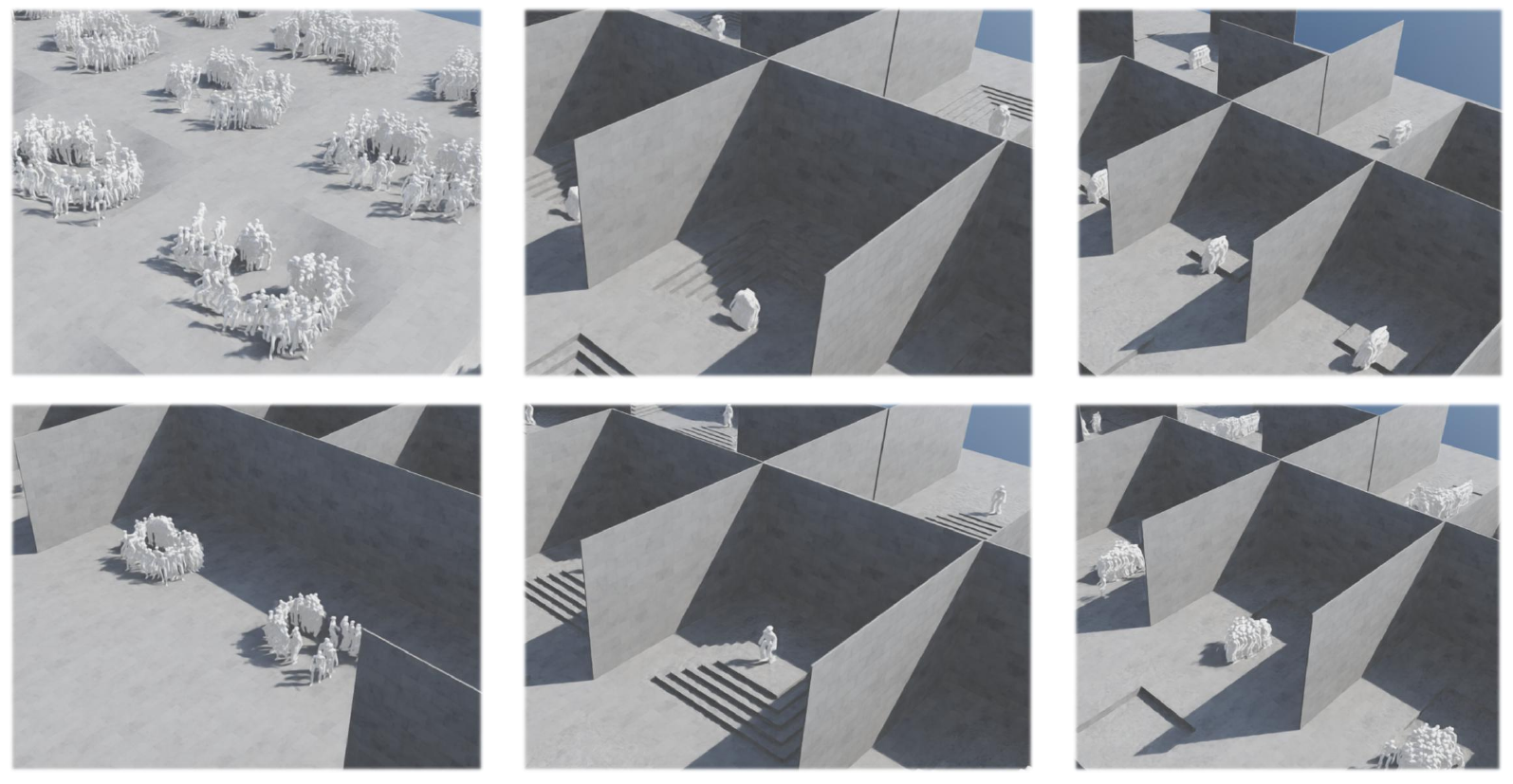}
    \caption{
    Simulation evaluation scenes across six terrain categories.
    Each policy is evaluated separately on flat ground, slopes, stair ascent, stair descent, platform step-up, and platform drop-down, with thousands of humanoid instances running in parallel.
    }
    \label{fig:sim_eval_scenes}
\end{figure*}

All simulation evaluations are conducted on a single NVIDIA RTX 4090 GPU with $4096$ parallel environments. 
For each evaluation run, we generate a terrain grid with $5$ rows and $10$ columns, and collect one episode from each environment. 
We report the mean of phase-specific metrics over the $4096$ evaluation episodes.

We evaluate each policy separately on six terrain categories: flat ground, slopes, upstairs, downstairs, up-platforms, and down-platforms. 
Each run contains only one terrain category, so contact metrics are averaged within the same terrain type. 
For fair comparison across policies, evaluations on the same terrain use cached terrain generation and a fixed random seed, ensuring that all policies are tested on identical terrain instances. Figure~\ref{fig:sim_eval_scenes} visualizes the parallel evaluation scenes, where thousands of humanoid instances are tested on the six terrain categories under the same simulation setup.

We do not report contact-area ratio or CoP margin for platform step-up and drop-down.
These metrics characterize sustained \textsc{Stance} support, whereas each platform trial contains
a single step transition and is designed primarily to evaluate the \textsc{Landing} event.
We therefore evaluate touchdown impact for these trials rather than steady stance support.

\subsection{Single-Critic Ablation}
\label{app:single_critic_ablation}

We further compare the proposed dual-critic design with a single-critic variant trained with the same reward terms and training budget.
Table~\ref{tab:app_single_critic} shows that the dual-critic policy consistently reduces touchdown force and improves stance support metrics on most terrains.
This suggests that separating dense and sparse reward returns helps the policy learn contact-quality objectives without sacrificing locomotion stability.

\begin{table*}[t]
\centering
\caption{
Single-critic ablation in simulation.
We compare the proposed dual-critic policy with a single-critic variant.
Lower is better for landing impact force; higher is better for CoP margin and contact area ratio.
Best values between the two variants are highlighted in bold.
}
\label{tab:app_single_critic}
\scriptsize
\setlength{\tabcolsep}{5.2pt}
\renewcommand{\arraystretch}{1.12}
\resizebox{\textwidth}{!}{
\begin{tabular}{l cc cc cc}
\toprule
\multirow{2}{*}{Terrain}
& \multicolumn{2}{c}{$F_{\mathrm{impact}}\downarrow$ [N]}
& \multicolumn{2}{c}{$d_{\mathrm{CoP}}\uparrow$ [mm]}
& \multicolumn{2}{c}{$A_c\uparrow$} \\
\cmidrule(lr){2-3}
\cmidrule(lr){4-5}
\cmidrule(lr){6-7}
& Dual critic & Single critic
& Dual critic & Single critic
& Dual critic & Single critic \\
\midrule
Flat
& \textbf{278.16} & 307.23
& \textbf{35.79} & 34.33
& \textbf{0.929} & 0.824 \\

Slope
& \textbf{297.32} & 335.90
& \textbf{34.24} & 31.40
& \textbf{0.805} & 0.635 \\

Stair descent
& \textbf{320.74} & 579.83
& \textbf{29.43} & 28.08
& \textbf{0.545} & 0.516 \\

Stair ascent
& \textbf{304.45} & 402.71
& \textbf{29.05} & 28.63
& \textbf{0.536} & 0.488 \\

Platform drop-down
& \textbf{310.62} & 333.93
& -- & --
& -- & -- \\

Platform step-up
& \textbf{310.73} & 421.35
& -- & --
& -- & -- \\
\bottomrule
\end{tabular}
}
\end{table*}

\section{Training Details}
\subsection{Reward Design}
\label{app:reward_design}

\paragraph{Reward list.}
We use a dual-critic reward decomposition based on the temporal structure of the reward signals.
For each reward group $g$, the reward manager computes
\begin{equation}
R_g(t)=\Delta t \sum_i w_i f_i(t),
\qquad \Delta t=0.02~\mathrm{s},
\end{equation}
where $w_i$ is the reward weight and $f_i(t)$ is the raw reward term.
The terms are divided into dense and sparse groups from the perspective of value learning, rather than by task semantics or by whether they are phase-conditioned.
The dense group contains rewards that provide high-frequency, continuous, and stable learning signals, including command tracking, posture regulation, smoothness penalties, and dense tactile shaping terms.
A phase-conditioned reward can still be dense if it is active at every step within its corresponding phase.
The sparse group contains rewards whose informative signals are intermittent, event-triggered, command-gated, or mainly activated by constraint violations.
Conversely, some sparse-group terms are evaluated every step but become informative only under specific events, violations, or command gates.
Each group is assigned to a separate critic, while the actor is updated using a mixed advantage from both value heads.
This decomposition reduces value-estimation interference between continuous locomotion shaping and sparse contact, task-completion, or safety objectives.

We denote the commanded base velocity by $\mathbf{c}=[c_x,c_y,c_{\omega_z}]$, the base-frame root linear and angular velocities by $\mathbf{v}_b$ and $\boldsymbol{\omega}_b$, the joint position and velocity by $\mathbf{q}$ and $\dot{\mathbf{q}}$, the default joint position by $\mathbf{q}^0$, the applied torque by $\boldsymbol{\tau}$, and the policy action by $\mathbf{a}_t$.
$\mathbbm{1}[\cdot]$ denotes the indicator function.
The ``Activation'' column describes when a term is evaluated or becomes informative; it is not the criterion used to define the reward group.

\begin{table*}[b]
\centering
\caption{
Dense reward group.
These rewards are optimized by the dense critic and provide frequent per-step learning signals, either throughout the episode or continuously within their active contact phase.
}
\label{tab:app_dense_rewards}
\scriptsize
\setlength{\tabcolsep}{2.8pt}
\renewcommand{\arraystretch}{1.15}
\resizebox{\textwidth}{!}{
\begin{tabular}{l c l l l}
\toprule
\textbf{Reward} & \textbf{Weight / scale} & \textbf{Formula} & \textbf{Activation} & \textbf{Purpose} \\
\midrule
$R_{\mathrm{lin}}$
& $2.0$
& $\exp\!\left(-\frac{(c_x-v_{b,x})^2+(c_y-v_{b,y})^2}{0.5^2}\right)$
& Per-step
& Track planar velocity \\

$R_{\mathrm{yaw}}$
& $2.0$
& $\exp\!\left(-\frac{(c_{\omega_z}-\omega_{b,z})^2}{0.5^2}\right)$
& Per-step
& Track yaw rate \\

$R_{\mathrm{yaw\_cmd}}$
& $-1.0$
& $|c_{\omega_z}|$
& Per-step
& Penalize excessive yaw-rate command magnitude \\

$R_{\mathrm{alive}}$
& $3.0$
& $\mathbbm{1}[\mathrm{not~terminated}]$
& Per-step
& Encourage stable rollouts \\

$R_{\mathrm{hip}}$
& $-0.5$
& $\sum_{j\in\mathcal{J}_{\mathrm{hip}}}(q_j-q^0_j)^2$
& Per-step
& Penalize hip yaw/roll deviation \\

$R_{\omega xy}$
& $-0.05$
& $\omega_{b,x}^2+\omega_{b,y}^2$
& Per-step
& Suppress base roll/pitch angular motion \\

$R_{\tau}$
& $-1.5{\times}10^{-7}$
& $\sum_{j\in\mathcal{J}_{\mathrm{lower}}}\tau_j^2$
& Per-step
& Penalize lower-body torque \\

$R_{\ddot{q}}$
& $-1.25{\times}10^{-7}$
& $\sum_j \ddot{q}_j^2$
& Per-step
& Penalize joint acceleration \\

$R_{\dot{q}}$
& $-1.0{\times}10^{-4}$
& $\sum_j \dot{q}_j^2$
& Per-step
& Penalize joint velocity \\

$R_{\Delta a}$
& $-0.006$
& $\|\mathbf{a}_t-\mathbf{a}_{t-1}\|^2$
& Per-step
& Encourage smooth actions \\

$R_{\mathrm{base\_ori}}$
& $-3.0$
& $g_{b,x}^2+g_{b,y}^2$
& Per-step
& Keep base upright \\

$R_{\mathrm{pelvis\_ori}}$
& $-3.0$
& $g_{\mathrm{pelvis},x}^2+g_{\mathrm{pelvis},y}^2$
& Per-step
& Keep pelvis upright \\

$R_{\mathrm{sw\_h}}$
& $0.12$
& $s_{\mathrm{warm}}\sum_k \alpha_{\mathrm{sw},k}\,0.12\,A_{\mathrm{cmd}}\operatorname{clip}(h^{\mathrm{eff}}_k/0.12,0,1)$
& Swing-gated dense
& Encourage swing-foot clearance \\

$R_{\mathrm{land\_dense}}$
& $0.15,0.12$
& $s_{\mathrm{warm}}\sum_k \alpha_{\mathrm{land},k}\left[-0.15e^2_{F,k}-0.12e^2_{\dot{F},k}\right]$
& Landing-window dense
& Provide dense shaping during touchdown \\

$R_{\mathrm{cop}}$
& $0.26$
& $s_{\mathrm{warm}}\sum_k \alpha_{\mathrm{st},k}\,0.26\,\operatorname{clip}(m_k/0.038,0,1)$
& Stance-gated dense
& Keep CoP inside the support region \\

$R_{\mathrm{area}}$
& $0.30$
& $s_{\mathrm{warm}}\sum_k \alpha_{\mathrm{st},k}\,0.30\,\operatorname{clip}(A_k,0,1)$
& Stance-gated dense
& Encourage larger contact area \\

$R_{\Delta\mathrm{cop}}$
& $0.03$
& $s_{\mathrm{warm}}\sum_k \alpha_{\mathrm{st},k}\,0.03\,\exp[-(\Delta\mathrm{CoP}_k/0.01)^2]$
& Stance-gated dense
& Encourage smooth CoP evolution \\

$R_{\mathrm{energy}}$
& $-5.0{\times}10^{-5}$
& $\sum_{j\in\mathcal{J}_{\mathrm{lower}}}\left(\frac{\tau_j\dot{q}_j}{k_j}\right)^2$
& Per-step
& Penalize normalized mechanical power \\

$R_{\mathrm{upper}}$
& $-0.004$
& $\sum_{j\in\mathcal{J}_{\mathrm{upper}}}|q_j-q^0_j|$
& Per-step
& Keep upper body near nominal posture \\
\bottomrule
\end{tabular}
}
\end{table*}

\begin{table*}[t]
\centering
\caption{
Sparse reward group.
These rewards are optimized by the sparse critic and become informative only under specific events, command gates, contact gates, phase gates, or constraint violations.
}
\label{tab:app_sparse_rewards}
\scriptsize
\setlength{\tabcolsep}{2.8pt}
\renewcommand{\arraystretch}{1.16}
\resizebox{\textwidth}{!}{
\begin{tabular}{l c l l l}
\toprule
\textbf{Reward} & \textbf{Weight / scale} & \textbf{Formula} & \textbf{Activation} & \textbf{Purpose} \\
\midrule
$R_{\mathrm{dont\_wait}}$
& $-0.5$
& $\mathbbm{1}[c_x>0.3]\!\left(\mathbbm{1}[v_{b,x}<0.15]+\mathbbm{1}[v_{b,x}<0]+\mathbbm{1}[v_{b,x}<-0.15]\right)$
& Command-gated
& Discourage stopping under forward commands \\

$R_{\mathrm{target}}$
& $0.1$
& $\mathbbm{1}\!\left[\|\mathbf{p}^{xy}_{\mathrm{target}}-\mathbf{p}^{xy}_{\mathrm{root}}\|\leq0.4\right]$
& Target event
& Reward reaching the target \\

$R_{\mathrm{stand}}$
& $-0.3$
& $\left(\sum_j |q_j-q^0_j|-4.0\right)\mathbbm{1}[\|\mathbf{c}_{xy}\|<0.15]\mathbbm{1}[|c_{\omega_z}|<0.15]$
& Command-gated
& Regulate standing posture \\

$R_{\mathrm{air}}$
& $0.6$
& $\min_k m_k$ if exactly one foot is in contact, else $0$
& Contact-pattern gated
& Encourage alternating swing and stance timing \\

$R_{\mathrm{slide}}$
& $-0.4$
& $\sum_k \mathbbm{1}\!\left[\max_{\mathrm{hist}}\|\mathbf{F}_k\|>1.0\right]\|\mathbf{v}^{w,xy}_{\mathrm{foot},k}\|$
& Contact-gated
& Penalize foot sliding during contact \\

$R_{\mathrm{plane}}$
& $-0.1$
& $\sum_{k\in\{L,R\}}\sum_r \operatorname{clip}(z_k-z_{\mathrm{scan},k,r}-0.058,0,0.3)\mathbbm{1}_k$
& Contact-gated
& Encourage the sole to stay close to local terrain \\

$R_{\mathrm{feet\_sep}}$
& $0.5$
& $\exp\!\left(-\frac{\max(0.20-d_{\mathrm{feet},y},0)}{0.05}\right)-1$
& Proximity-gated
& Penalize excessively close foot placement \\

$R_{\mathrm{pre\_v}}$
& $0.30$
& $s_{\mathrm{warm}}\sum_k \alpha_{\mathrm{pre},k}\,0.30[-\max(-v^z_{\mathrm{foot},k}-0.24,0)^2]$
& Pre-landing gated
& Reduce excessive downward velocity before touchdown \\

$R_{\mathrm{pre\_a}}$
& $0.08$
& $s_{\mathrm{warm}}\sum_k \alpha_{\mathrm{pre},k}\,0.08[-\max(-a^z_{\mathrm{foot},k}-6.0,0)^2]$
& Pre-landing gated
& Reduce excessive downward acceleration before touchdown \\

$R_{\mathrm{land\_F}}$
& $0.44$
& $s_{\mathrm{warm}}\sum_k \mathbbm{1}^{k}_{\mathrm{land\_end}}(-0.44\,F^{\mathrm{peak}}_k/W)$
& Landing event
& Penalize touchdown peak normal force \\

$R_{\mathrm{land\_dF}}$
& $0.16$
& $s_{\mathrm{warm}}\sum_k \mathbbm{1}^{k}_{\mathrm{land\_end}}(-0.16\,\dot{F}^{\mathrm{peak},+}_k/40000)$
& Landing event
& Penalize touchdown force-rate peak \\

$R_{\mathrm{land\_\rho}}$
& $0.12$
& $s_{\mathrm{warm}}\sum_k \mathbbm{1}^{k}_{\mathrm{land\_end}}(-0.12\,\rho^{\mathrm{peak}}_k)$
& Landing event
& Penalize concentrated point loading \\

$R_{\mathrm{pos\_lim}}$
& $-1.0$
& $\sum_j \left(\max(q^{\min}_j-q_j,0)+\max(q_j-q^{\max}_j,0)\right)$
& Violation-gated
& Penalize joint position-limit violation \\

$R_{\mathrm{vel\_lim}}$
& $-1.0$
& $\sum_j \operatorname{clip}(|\dot{q}_j|-0.9\dot{q}^{\max}_j,0,1)$
& Violation-gated
& Penalize joint velocity-limit violation \\

$R_{\tau\_\mathrm{lim}}$
& $-0.1$
& $\sum_j \max(|\tau_j|-0.8\tau^{\max}_j,0)^2$
& Violation-gated
& Penalize torque-limit violation \\

$R_{\mathrm{bad\_contact}}$
& $-1.0$
& $\sum_b \mathbbm{1}\!\left[\max_{\mathrm{hist}}\|\mathbf{F}_b\|>1.0\right]$
& Contact-event
& Penalize undesired body contacts \\
\bottomrule
\end{tabular}
}
\end{table*}

For phase-aware rewards, $\alpha_{\mathrm{sw},k}$, $\alpha_{\mathrm{pre},k}$, $\alpha_{\mathrm{land},k}$, and $\alpha_{\mathrm{st},k}$ denote the eligibility of foot $k$ for \textsc{Swing}, \textsc{PreLanding}, \textsc{Landing}, and \textsc{Stance}, respectively.
The warmup coefficient is
\begin{equation}
s_{\mathrm{warm}}=\min\left(\frac{\mathrm{iteration}}{5000},1\right).
\end{equation}
The command-activity gate is
\begin{equation}
A_{\mathrm{cmd}}=
\mathbbm{1}[\|\mathbf{c}_{xy}\|>0.15~\mathrm{or}~|c_{\omega_z}|>0.15].
\end{equation}
$h^{\mathrm{eff}}_k$ is the raycast-based effective foot height, $F_k$ is the tactile normal force, $W$ is the robot body weight used for normalization, $A_k$ is the contact-area ratio, and $m_k$ is the signed CoP margin inside the tactile foot polygon.
For the dense landing term,
\begin{equation}
e_{F,k}=\max(F_k/W-1,0),
\qquad
e_{\dot{F},k}=\max(\dot{F}^{+}_k/40000-1,0),
\end{equation}
where $\dot{F}^{+}_k=\max(\dot{F}_k,0)$.
The touchdown-event terms are applied at the end of the landing window using peak statistics, while $R_{\mathrm{land\_dense}}$ provides per-step shaping during the landing phase.

\paragraph{AMP auxiliary reward.}
In addition to the environment rewards, we use an AMP auxiliary reward for motion regularization.
This reward is applied to the overall policy objective rather than being tied to a single reward group:
\begin{equation}
r_{\mathrm{AMP}}
=
0.25\,\max\left(1-0.25(D(s)-1)^2,0\right),
\end{equation}
where $D(s)$ is the discriminator output for the policy AMP state.
The AMP term encourages a natural motion style and is optimized together with the dense and sparse environment rewards.

\subsection{Terrain Curriculum}
\label{app:terrain_curriculum}

We use a terrain curriculum to expose the policy to progressively harder foot--terrain interactions while maintaining stable exploration.
At the beginning of training, a fixed terrain grid with $10$ difficulty rows and $20$ terrain columns is generated.
Each row corresponds to a difficulty level, and each column corresponds to one terrain category.
During training, the terrain geometry is kept fixed; the curriculum only updates the terrain origin assigned to each environment.

For row $r$, the terrain difficulty is sampled as
\begin{equation}
d=\frac{r+\eta}{10}, \qquad \eta\sim\mathcal{U}(0,1),
\end{equation}
where lower rows contain easier terrains and higher rows contain harder terrains.
Environments are initialized from low-to-medium difficulty levels and are promoted or demoted according to their episode-level velocity-tracking performance.
This curriculum allows the robot to first acquire stable locomotion and then gradually face stronger contact disturbances, larger height transitions, and more demanding support conditions.

Table~\ref{tab:app_terrain_curriculum} summarizes the terrain categories used for training.
The rough-flat terrains mainly improve stance robustness and contact stability, while the stair and platform terrains target swing clearance, touchdown regulation, and support quality under discrete height changes.

\begin{table*}[h]
\centering
\caption{
Terrain curriculum configuration.
The curriculum parameter is interpolated from the easiest to the hardest row.
}
\label{tab:app_terrain_curriculum}
\scriptsize
\setlength{\tabcolsep}{4.0pt}
\renewcommand{\arraystretch}{1.16}
\resizebox{\textwidth}{!}{
\begin{tabular}{l l l l}
\toprule
\textbf{Terrain category} & \textbf{Task setting} & \textbf{Curriculum parameter} & \textbf{Purpose} \\
\midrule
Rough flat
& Forward locomotion on uneven ground
& Perlin height scale: $0.00$--$0.02$
& Robust walking under mild contact-height perturbations \\

Rough standing
& Standing on uneven ground
& Perlin height scale: $0.00$--$0.05$
& Static support stability and CoP regulation \\

Stair descent
& Walking down stairs
& Step height: $0.05$--$0.23\,\mathrm{m}$
& Soft touchdown and impact reduction during downward transitions \\

Stair ascent
& Walking up stairs
& Step height: $0.05$--$0.23\,\mathrm{m}$
& Swing clearance and foothold establishment during upward transitions \\

Platform drop-down
& Stepping down from high platforms
& Platform height: $0.05$--$0.40\,\mathrm{m}$
& High-impact landing control and stance recovery \\

Platform step-up
& Stepping onto high platforms
& Platform height: $0.05$--$0.40\,\mathrm{m}$
& Large-step clearance, foot placement, and push-off coordination \\
\bottomrule
\end{tabular}
}
\end{table*}

\subsection{Training Hyperparameters}
\label{app:training_hyperparameters}

All policies are trained with the same budget unless otherwise specified: $2048$ parallel environments for $50{,}000$ iterations.
Each PPO iteration collects $24$ steps per environment, resulting in $49{,}152$ transitions per update.
We use a dual-critic AMP-PPO implementation, where the actor receives a mixed advantage from the locomotion and contact-quality value heads.
Table~\ref{tab:app_training_hyperparameters} summarizes the main training hyperparameters.

\begin{table*}[t]
\centering
\caption{
Training hyperparameters.
All baselines and ablations use the same PPO budget and optimization settings unless otherwise stated.
}
\label{tab:app_training_hyperparameters}
\scriptsize
\setlength{\tabcolsep}{4.0pt}
\renewcommand{\arraystretch}{1.08}
\resizebox{\textwidth}{!}{
\begin{tabular}{l p{0.78\textwidth}}
\toprule
\textbf{Category} & \textbf{Configuration} \\
\midrule
Environment
& $2048$ environments; $24$ rollout steps per environment; $49{,}152$ transitions per update; $50{,}000$ training iterations. \\

Policy
& Encoder MoE actor-critic with $4$ experts; actor MLP $[256,128,64]$; critic MLP $[256,128,64]$; ELU activation; initial action std. $1.0$. \\

Depth encoder
& CNN output dimension $128$; channels $[4]$; kernel/stride/padding $3/1/1$; MLP hidden dimensions $[256,256]$. \\

PPO
& AdamW optimizer; learning rate $1.0\times10^{-3}$ with adaptive KL schedule; desired KL $0.01$; $4$ mini-batches; mini-batch size $12{,}288$; $5$ epochs per update; clip range $0.2$; $\gamma=0.99$; $\lambda=0.95$; entropy coefficient $0.006$; value loss coefficient $1.0$; max gradient norm $1.0$. \\

Dual critic
& Two value heads estimate dense and sparse environment returns, respectively; advantage mixing weights $[1.1,0.9]$. \\

AMP
& Discriminator MLP $[1024,512]$ with ReLU; AdamW optimizer with learning rate $1.0\times10^{-4}$; discriminator reward coefficient $0.25$; gradient penalty coefficient $5.0$; weight decay $3.0\times10^{-4}$; logit decay $0.04$. \\
\bottomrule
\end{tabular}
}
\end{table*}

The two critic heads estimate returns for the dense and sparse reward groups, respectively.
The AMP auxiliary reward is used as an additional motion-regularization objective and is optimized together with both reward groups.

\subsection{Domain Randomization}
\label{app:domain_randomization}

We apply domain randomization to physical parameters, actuation, initial states, external disturbances, observations, exteroceptive perception, tactile sensing, and command generation.
Uniform sampling is denoted by $\mathcal{U}(\cdot,\cdot)$ and log-uniform scaling by $\log\mathcal{U}(\cdot,\cdot)$.
The active randomization terms are summarized in Table~\ref{tab:app_domain_randomization}.

\begin{table*}[t]
\centering
\caption{
Domain randomization and training-distribution randomization.
The actor observes corrupted sensing inputs, while the critic uses clean privileged observations during training.
}
\label{tab:app_domain_randomization}
\scriptsize
\setlength{\tabcolsep}{3.4pt}
\renewcommand{\arraystretch}{1.13}
\resizebox{\textwidth}{!}{
\begin{tabular}{l l l}
\toprule
\textbf{Category} & \textbf{Randomized quantity} & \textbf{Range / distribution} \\
\midrule
\multirow{3}{*}{Contact material}
& Static friction & $\mathcal{U}(0.3,1.6)$ \\
& Dynamic friction & $\mathcal{U}(0.3,1.6)$, with $\mu_d\leq\mu_s$ \\
& Restitution & $\mathcal{U}(0.05,0.5)$ \\
\midrule

\multirow{4}{*}{Robot model}
& Default joint-position offset & $\mathcal{U}(-0.01,0.01)\,\mathrm{rad}$ \\
& Torso CoM offset, $x$ & $\mathcal{U}(-0.025,0.025)\,\mathrm{m}$ \\
& Torso CoM offset, $y,z$ & $\mathcal{U}(-0.05,0.05)\,\mathrm{m}$ \\
& Torso mass offset & $\mathcal{U}(-2.0,2.0)\,\mathrm{kg}$ \\
\midrule

\multirow{3}{*}{Actuation}
& PD stiffness scale & $\log\mathcal{U}(0.95,1.05)$ \\
& PD damping scale & $\log\mathcal{U}(0.95,1.05)$ \\
& Actuator delay & $\mathcal{U}\{0,1,2\}$ control steps \\
\midrule

\multirow{5}{*}{Reset state}
& Root position, $x,y$ & $\mathcal{U}(-0.1,0.1)\,\mathrm{m}$ \\
& Root yaw & $\mathcal{U}(-0.1,0.1)\,\mathrm{rad}$ \\
& Root linear velocity & $\mathcal{U}(-0.2,0.2)\,\mathrm{m/s}$ \\
& Root angular velocity & $\mathcal{U}(-0.2,0.2)\,\mathrm{rad/s}$ \\
& Joint position & $\mathcal{U}(-0.15,0.15)\,\mathrm{rad}$ \\
\midrule

\multirow{3}{*}{External disturbance}
& Push interval & $\mathcal{U}(7.0,10.0)\,\mathrm{s}$ \\
& Root velocity impulse, $x,y$ & $\mathcal{U}(-0.5,0.5)\,\mathrm{m/s}$ \\
& Root angular-velocity impulse, roll/pitch/yaw & $\mathcal{U}(-0.52,0.52)$, $\mathcal{U}(-0.52,0.52)$, $\mathcal{U}(-0.78,0.78)\,\mathrm{rad/s}$ \\
\midrule

\multirow{4}{*}{Actor observation}
& Base angular velocity noise & $\mathcal{U}(-0.2,0.2)$ \\
& Projected gravity noise & $\mathcal{U}(-0.05,0.05)$ \\
& Joint position noise & $\mathcal{U}(-0.01,0.01)$ \\
& Joint velocity noise & $\mathcal{U}(-0.5,0.5)$ \\
\midrule

\multirow{2}{*}{Camera}
& Position offset & $x:\mathcal{U}(-0.003,0.003)$, $y:\mathcal{U}(-0.01,0.01)$, $z:\mathcal{U}(-0.005,0.005)\,\mathrm{m}$ \\
& Orientation offset & roll: $\mathcal{U}(-0.02,0.02)$, pitch/yaw: $\mathcal{U}(-0.03,0.03)\,\mathrm{rad}$ \\
\midrule

\multirow{4}{*}{Depth}
& Range-based Gaussian noise & $\sigma=0.04$ for depths in $[0.5,2.5]\,\mathrm{m}$ \\
& Artifact probability & $10^{-4}$ \\
& Gaussian blur & kernel size $3$, $\sigma=1$ \\
& Random image noise & probability $0.1$, $\sigma=0.2$ \\
\midrule

\multirow{3}{*}{Tactile}
& Relative taxel force error & $\mathcal{U}(0.90,1.10)$ \\
& Taxel XY offset & $\operatorname{clip}(\mathcal{N}(0,0.0015^2),-0.005,0.005)\,\mathrm{m}$ \\
& Measurement delay & probability $0.10$, maximum delay $1$ frame \\
\midrule

\multirow{4}{*}{Command}
& Forward velocity & $\mathcal{U}(0.45,0.55)\,\mathrm{m/s}$ \\
& Lateral velocity & $0$ \\
& Yaw rate & $\mathcal{U}(-1.0,1.0)\,\mathrm{rad/s}$ \\
& Standing environments & $5\%$ \\
\bottomrule
\end{tabular}
}
\end{table*}

Observation corruption is applied to the actor but not to the critic.
This keeps value estimation stable while exposing the deployed policy to realistic proprioceptive, visual, and tactile uncertainty.

\section{Tactile Sensing and Feature Processing}
\subsection{Tactile Noise Modeling in Simulation}
\label{app:tactile_noise}

To improve robustness to sensing uncertainty, we perturb the simulated tactile measurements during training.
The noise is applied to the taxel-level force map before computing the compact tactile features used by the policy.
It models three common deployment errors: force-scale uncertainty, taxel-coordinate calibration error, and short measurement delay.
Table~\ref{tab:app_tactile_noise} summarizes the configuration.

\begin{table}[t]
\centering
\caption{
Tactile measurement noise used during training.
The perturbations are applied before extracting the compact tactile features.
}
\label{tab:app_tactile_noise}
\scriptsize
\setlength{\tabcolsep}{4.0pt}
\renewcommand{\arraystretch}{1.12}
\begin{tabular}{l c l}
\toprule
\textbf{Noise source} & \textbf{Value} & \textbf{Description} \\
\midrule
Relative taxel force error 
& $\pm 10\%$
& Per-taxel scaling $s_i\sim\mathcal{U}(0.9,1.1)$ \\

Taxel XY offset std.
& $0.0015\,\mathrm{m}$
& Episode-level taxel-coordinate perturbation \\

Taxel XY offset clip
& $0.005\,\mathrm{m}$
& Maximum coordinate perturbation \\

Delay probability
& $0.08$
& Probability of using delayed tactile history \\

Maximum delay
& $1$ frame
& At most one tactile update delay \\

Force clipping
& $F_i\geq0$
& Clamp measured taxel force to be nonnegative \\

Dropout / quantization
& Disabled
& Not used in the current configuration \\
\bottomrule
\end{tabular}
\end{table}

The taxel-coordinate perturbation is sampled once per episode and remains fixed within the episode, modeling calibration and mounting offsets rather than frame-wise noise.
The delayed measurement is sampled per environment and per foot, and is applied to the full taxel vector of that foot.
The noisy taxel forces are not renormalized to the clean total force after perturbation, allowing the resulting normal force, contact area, and CoP features to vary under tactile measurement uncertainty.

\subsection{Feature Normalization}
\label{app:tactile_normalization}

The actor observes a compact tactile state for each foot, consisting of contact area ratio, normalized normal force, and normalized CoP.
The full taxel map is not provided to the policy.
For each foot $f\in\{L,R\}$, the actor tactile feature is
\begin{equation}
\mathbf{x}^{f}_{\mathrm{tac}}
=
[A^f,\tilde{F}^f,\tilde{p}^{f}_x,\tilde{p}^{f}_y].
\end{equation}
Here $A^f$ is the contact area ratio clipped to $[0,1]$.
The normal force feature is computed from the sum of the noisy taxel forces and normalized by the robot body weight:
\begin{equation}
\tilde{F}^f
=
\operatorname{clip}
\left(
\frac{\sum_i \hat{F}^{f}_i}{W},
0,1.5
\right),
\qquad
W=mg,
\end{equation}
where $\hat{F}^{f}_i$ denotes the noisy measured force at taxel $i$.
The CoP coordinates are normalized by the positive and negative extents of the foot outline in the ankle-roll frame and clipped to $[-1,1]$.
When $\tilde{F}^f<0.05$, the normalized CoP is set to zero to avoid exposing unreliable near-zero-contact estimates to the policy.

We use a two-frame tactile history for the actor.
Thus, the tactile observation contributes $2\times2\times4=16$ dimensions.
The critic receives the same tactile features and additionally uses the normalized vertical foot velocity during training.
The deployed actor does not use this velocity term, keeping its tactile observation consistent with real hardware sensing.

\subsection{Insole Sampling Rate and Peak-Force Validation}

The wireless tactile insole used for all hardware evaluations operates at 25 Hz. 
Once a sample is acquired, post-acquisition processing and communication introduce less than 
1 ms of additional latency; wired acquisition supports up to 100 Hz. 
Because finite sampling may underestimate an instantaneous touchdown peak, we repeated hardware 
tests at 100 Hz on flat ground and platform drop-down. The measured impact forces were 
$199.2 \pm 14.9$ N and $427.7 \pm 20.1$ N, respectively, compared with 
$191.3 \pm 17.7$ N and $404.6 \pm 43.2$ N under the 25-Hz protocol. 
The agreement indicates that the touchdown rising-edge measurement provides a consistent estimate 
of impact magnitude for our evaluation. All policy comparisons use the same wireless sensing and 
processing pipeline.

\end{document}